\documentclass[letterpaper, 10 pt, conference]{ieeeconf}  

\IEEEoverridecommandlockouts                              

\usepackage{algorithm}
\usepackage{algpseudocode}
\usepackage{algorithmicx}
\usepackage{amsmath}
\usepackage{amssymb}
\usepackage{subfig}
\usepackage[table]{xcolor} 

\usepackage{multirow}
\usepackage{newfloat}
\usepackage{listings}
\usepackage{cite}
\usepackage{graphicx}

\usepackage{gensymb}

\usepackage{booktabs}

\title{\LARGE \bf
Contextual Neural Moving Horizon Estimation \\for Robust Quadrotor Control in Varying Conditions
}

\author{Kasra Torshizi*$^{1}$, Chak Lam Shek*$^{1}$, Khuzema Habib$^{1}$, Guangyao Shi$^{2}$, Pratap Tokekar$^{1}$, Troi Williams$^{1}$
\thanks{}
\thanks{* Equal Contribution}
\thanks{$^{1}$
       University of Maryland, College Park. {\tt\small \{ktorsh, cshek1, khabib, tokekar, troiw\}@umd.edu}.} 
\thanks{$^{2}$ University of Southern California. {\tt\small shig@usc.edu}%
}}

\begin{document}

\maketitle
\thispagestyle{empty}
\pagestyle{empty}

\begin{abstract}
Adaptive controllers on quadrotors typically rely on estimation of disturbances to ensure robust trajectory tracking. Estimating disturbances across diverse environmental contexts is challenging due to the inherent variability and uncertainty in the real world. Such estimators require extensive fine-tuning for a specific scenario, which makes them inflexible and brittle to changing conditions. Machine-learning approaches, such as training a neural network to tune the estimator’s parameters, are promising. However, collecting data across all possible environmental contexts is impossible. It is also inefficient as the same estimator parameters could work for ``nearby" contexts. In this paper, we present a sequential decision making strategy that decides which environmental contexts, using Bayesian Optimization with a Gaussian Process, to collect data from in order to ensure robust performance across a wide range of contexts. Our method, Contextual NeuroMHE, eliminates the need for exhaustive training across all environments while maintaining robust performance under different conditions. By enabling the neural network to adapt its parameters dynamically, our method improves both efficiency and generalization. Experimental results in various real-world settings demonstrate that our approach outperforms the prior work by 20.3\% in terms of maximum absolute position error and can capture the variations in the environment  with a few carefully chosen contexts.

\end{abstract}

\section{Introduction}
\begin{figure}
    \centering
    \includegraphics[width=1\linewidth]{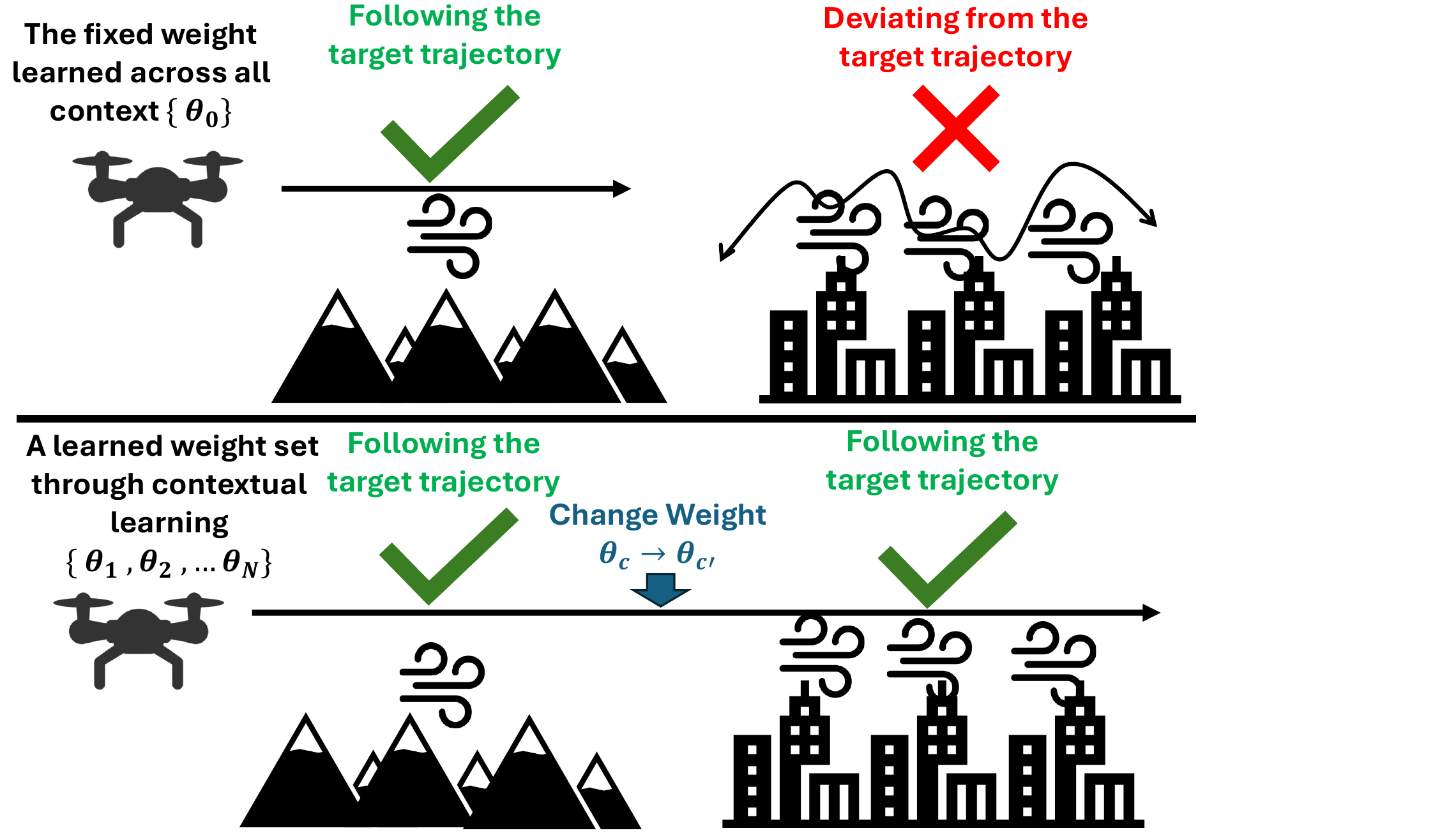}
    \caption{(top) Without adaptation, a single global model using fixed mapping $ \theta_0(\cdot) $ fails to generalize across varying environment contexts $ c, c' $, leading to deviation from the target trajectory. (bottom) In contrast, our approach learns a set of context-specific mappings $ \{ \theta_1(\cdot), \theta_2(\cdot), \ldots, \theta_N(\cdot) \} $ through contextual learning. When the context shifts from $ c $ to $ c' $, the model dynamically adapts by switching from $ \theta_c(\cdot) $ to $ \theta_{c'}(\cdot) $, enabling robust trajectory tracking under different environmental conditions.
    }
    \label{fig:1}
\end{figure}

Quadrotors have found widespread use in applications such as environmental monitoring~\cite{charrow2015information, zhu2021online}, precision agriculture~\cite{albani2017field,7989676}, and search and rescue operations~\cite{best2022resilient,8206510}. However, their deployment in diverse, dynamic environments presents a significant challenge. Quadrotors must adapt to unpredictable disturbances, such as fluctuating wind conditions and environmental interactions~\cite{Sikkel2016ANO, cioffi2023hdvioimprovinglocalizationdisturbance, Huang2023DATTDA}. To ensure reliable trajectory tracking and maintain control performance, it is essential that the controller~\cite{lee2011controlcomplexmaneuversquadrotor} explicitly accounts for disturbances through accurate disturbance estimation. Developing a universal model to estimate the full spectrum of disturbances encountered in various flight scenarios is impractical due to time and resource constraints~\cite{hanover2021performance}. Given the wide range of possible conditions, it is not feasible to train a separate model for each scenario. This underscores the need for adaptive methods capable of estimating disturbances in real-time and adjusting flight control parameters dynamically to handle changing environmental conditions (Figure \ref{fig:1}).

Disturbance estimation for quadrotors has been a long-standing area of research due to its critical importance in maintaining stable flight performance. Traditional approaches have largely relied on momentum-based estimators~\cite{6943154,6878116,6907146}, often implemented with stable low-pass filters. While such filters are effective for static or slowly varying disturbances, they struggle to capture rapidly changing forces. To overcome these limitations, more advanced methods such as Moving Horizon Estimation (MHE)~\cite{GIRRBACH2018116, robertson1996moving} and deep neural networks (DNNs)~\cite{article} have been investigated. MHE-based approaches formulate disturbance dynamics within a finite-horizon constrained optimization problem and have been widely adopted for state estimation. Although they demonstrate improved performance, these methods remain highly sensitive to the choice of internal parameters. To mitigate this, NeuroMHE~\cite{wang2023neural} has been introduced, which adaptively tunes estimator parameters based on the observed state. However, NeuroMHE is restricted to operation under a fixed disturbance distribution, limiting its applicability in diverse scenarios. Training multiple specialized models to cover a wide range of conditions is possible but incurs prohibitive data and computational demands.

Alternatively, recent research has increasingly focused on training smaller models tailored to individual tasks, receiving growing attention for their ability to capture context-specific dynamics. This form of contextual learning leverages environmental information to construct models specialized for narrow operating regimes, where data availability is often limited. For quadrotors, this adaptability is crucial for real-time adjustment to disturbances such as wind. Recent studies have demonstrated the efficacy of contextual learning in robotics and autonomous systems~\cite{Hu2025CARoL, Evans2022ContextAdaptation, Nagabandi2018MetaRL, 10802075}. Lee et al.~\cite{Lee2020ContextAwareDynamics} introduced a context-aware dynamics model that enhances generalization in model-based reinforcement learning, allowing robots to adapt to new, unseen conditions without retraining the entire system. Similarly, Huang et al.~\cite{Huang2025GeneralizationRL} developed a robust adaptation module that generalizes well in both in-distribution and out-of-distribution environments, effectively handling disturbances like wind and payload variations in real-time. 

However, many existing methods, including contextual learning methods, for disturbance estimation are tailored to specific scenarios and therefore fail to generalize across diverse environments. Training a single model that can estimate disturbances accurately under all conditions is often inefficient and unreliable—not only because it requires exclusive, environment-specific data, but also because it is fundamentally impossible to collect data covering the full spectrum of possible contexts. Moreover, even with large datasets, it is inevitable that certain contexts will remain underrepresented, leading to gaps that prevent robust generalization. This fundamental limitation underscores the need for approaches that can adaptively leverage informative contexts rather than attempting to exhaustively cover every possible environment.

To address these challenges, we propose an active learning method, Contextual NeuroMHE, for disturbance estimation that adapts to varying flight scenarios under resource constraints. Rather than training a separate model for each situation, we formulate the problem as a sequential decision task, where a Gaussian Process~\cite{rasmussen2006gaussian} using Bayesian Optimization~\cite{mockus1978bayesian} guides the selection of the most informative contexts that improve the estimator. This formulation leverages the property that as the number of training samples grows, the marginal gain in performance diminishes, making it essential to prioritize the most valuable contexts instead of collecting data randomly. By exploiting this principle, our method optimizes disturbance estimation with a limited budget, training a compact set of models that generalizes across diverse scenarios. This provides a scalable solution for real-time disturbance estimation and robust flight control across multiple environments.

The main contributions of this work are threefold:
\begin{enumerate}
    \item We formulate context selection for disturbance-aware estimation as a 
    {sequential decision problem}, enabling adaptive selection of informative training contexts.
    \item We propose a {novel learning framework} that integrates 
    GP–based context selection with NeuroMHE, allowing weight matrices 
    to adapt dynamically to varying environmental conditions.
    \item Through extensive real-world experiments on a Crazyflie 2.1 with household fans, 
covering 13 wind contexts across three environments, we demonstrate a \textbf{20.3\%} improvement 
in estimation and control performance over baseline approaches.
\end{enumerate}

\section{RELATED WORK}
Handling disturbances in quadrotor flight has long motivated the use of MHE, since conventional observers, such as Extended Kalman Filter (EKF)~\cite{julier2004unscented} often struggle with constraints, strong nonlinearities, and time-varying dynamics. Early works established MHE~\cite{GIRRBACH2018116, robertson1996moving} as a constrained optimization framework for state estimation, later extended to nonlinear systems with stability guarantees~\cite{RAO20011619, ALESSANDRI20081753}. Comparisons~\cite{doi:10.1021/ie034308l} showed MHE’s advantages over EKF under process noise and constraints, while algorithmic contributions~\cite{Tenny2002EfficientMH, zavala2008fast} focused on real-time feasibility through efficient optimization and sensitivity updates. Broader surveys~\cite{RAWLINGS20061529} positioned MHE within modern estimation alongside particle filtering, and more recent robotics-oriented work~\cite{5654354} demonstrated its utility for constrained robot localization. Building on these foundations, recent research has advanced MHE with data-driven and learning-based methods: differentiable MHE formulations for robust quadrotor flight control~\cite{wang2021differentiable, wang2023neural}, trust-region neural MHE that leverages second-order optimization for efficient tuning~\cite{10611703}, and Gaussian process-aided MHE to capture aerodynamic effects in agile quadrotor flight~\cite{10342084}. Collectively, the studies demonstrate that MHE is an effective framework for \textit{quadrotor disturbance estimation}, offering robustness to wind and environmental interactions and benefiting from recent advances in machine learning.

While prior approaches have shown success in specific settings, they often lack generality, requiring extensive case-specific training and manual tuning, while deep learning methods remain resource-intensive. To address these limitations, we propose a contextual method that adapts to diverse flight scenarios under limited data and computational budgets, enabling efficient and scalable disturbance estimation for robust quadrotor control.

\section{Preliminaries}

MHE is an optimization-based state estimation framework that computes the system state by solving a constrained optimization problem over a finite horizon of past measurements and control inputs. Unlike recursive estimators such as the Kalman filter, which update the state estimate based only on the most recent observation, MHE maintains a sliding window of information. This enables it to explicitly account for nonlinear dynamics, state and input constraints, and non-Gaussian disturbances. These properties make MHE particularly attractive for safety-critical robotic systems such as quadrotors, where robustness to disturbances is essential.

At each time step $t \geq N$, the MHE procedure estimates a sequence of states $\{x_k\}_{k=t-N}^{t}$ and process noise vectors $\{w_k\}_{k=t-N}^{t-1}$ by solving a nonlinear optimization problem of the form
\begin{equation}
\begin{aligned}
\min_{x, w} J = 
&\ \frac{1}{2} \|x_{t-N} - \hat{x}_{t-N}\|^2_P +
\frac{1}{2} \sum_{k=t-N}^{t} \|y_k - h(x_k)\|^2_{R_k} \\
&\ + \frac{1}{2} \sum_{k=t-N}^{t-1} \|w_k\|^2_{Q_k}
\end{aligned}
\end{equation}
subject to the discrete-time system dynamics
\begin{equation}
x_{k+1} = f(x_k, u_k, w_k, \Delta t),
\end{equation}
where $f(\cdot)$ denotes the dynamics model and $\Delta t$ is the sampling interval.
The objective consists of three components: (i) an arrival cost $||x_{t-N} - \hat{x}_{t-N}||^2_P$ that incorporates the prior estimate $\hat{x}_{t-N}$, (ii) a measurement cost penalizing the discrepancy between predicted and observed outputs, and (iii) a process cost penalizing deviations from the nominal dynamics. Here, $||x||^2_M = x^\top M x$ denotes the weighted norm induced by a positive-definite matrix $M$.

MHE performance is highly sensitive to these weights. If an improper set of weights is used, the resulting estimates $\hat{x}$ may be biased or inaccurate, compromising closed-loop performance. To mitigate this issue, NeuroMHE~\cite{wang2023neural} employs neural networks to learn weightings directly from data.

The parameter vector $\theta$ collects all tunable elements in the NeuroMHE weighting matrices: 

\begin{equation}
\theta = \{\text{vec}(P), \{ \text{vec}(R_k) \}_{k=t-N}^t, \{ \text{vec}(Q_k) \}_{t-N}^{t-1} \} \in \mathbb{R}^p,
\end{equation}
where $P$, $R_k$, and $Q_k$ are symmetric positive-definite matrices weighting the initial-state deviation, measurement residuals, and process noise, respectively.

A neural network outputs the weighting matrices or residual models, and analytic gradients of the MHE solution with respect to these weights are derived, enabling efficient embedding of MHE within a differentiable learning framework. NeuroMHE has shown the capability to automatically tune key MHE parameters across varying flight conditions, improving estimation accuracy without requiring ground-truth disturbance data.

In our formulation, a neural network parameterized by $\varpi$ maps input features to the elements of $\theta$. The network is trained end-to-end by minimizing a downstream control-oriented loss, such as trajectory tracking error. We define the context $c$ as a set of latent or observable variables that characterize the environment, such as wind conditions, sensor noise levels, or payload configurations. Importantly, within the same context $c$, the noise and disturbance statistics are assumed to follow a consistent Gaussian distribution with fixed parameters. This assumption ensures that models trained under a given context capture stable statistical properties, while variations across different contexts reflect shifts in the underlying distribution. By leveraging this structure, the neural network can exploit context-dependent regularities to produce appropriate weightings $\theta(c)$ that remain valid within each regime.

\section{Problem Formulation}
The objective of this paper is to adaptively tune the weighting matrices in MHE to minimize state estimation errors in downstream control tasks. Specifically, we define a scalar loss function $L(\hat{x}(\theta (c))$ where $\hat{x}$ denotes the estimated state produced by MHE, $c \in \mathcal{C} \subseteq \mathbb{R}^d$ represents a continuous context variable that modulates the noise statistics, and $\theta$ denotes the weights of the MHE. This loss function is designed to penalize trajectory tracking errors or other task-relevant quantities, enabling end-to-end optimization of the estimation process with respect to context.

A key challenge is that collecting data across the full spectrum of contexts $\mathcal{C}$ is infeasible. In practice, training can only rely on a finite set of sampled contexts, which may not fully capture rare or extreme operating regimes. This limitation motivates methods that can generalize from limited training data by actively and selectively choosing training contexts that maximize overall performance.

To enable such adaptation, we formulate the problem as a sequential decision problem, where the learner must iteratively select the most informative contexts and update the associated weighting functions. We model this as learning a set of $M$ neural network mappings ${\boldsymbol{\theta} = \{\theta_{\varpi_{c_1}}(\cdot), \theta_{\varpi_{c_2}}(\cdot), \ldots, \theta_{\varpi_{c_M}}(\cdot)\}}$, where each $\theta_{\varpi_{c_i}}(\cdot)$ takes the historical state associated with context $c_i$ as input and dynamically outputs the optimal weighting parameters for the MHE. Here, $\varpi$ denotes the parameters of the neural networks.

Formally, the sequential nature of the problem can be expressed using recursive evaluations of performance. Based on previous losses ${L(\theta_{\varpi_{c_1}}, c), \ldots, L(\theta_{\varpi_{c_{k-1}}}, c)}$ for all $c \in \mathcal{C}$, we define 
\begin{equation}
\begin{aligned}L(\theta_{\varpi_{c_k}}, c'; \theta_{\varpi_{c_{1:k-1}}}) = \min \big(L(\theta_{\varpi_{c_k}}, c'), L(\theta_{\varpi_{c_{1:k-1}}}, c')\big), \\
\quad \forall c' \in \mathcal{C} ; k > 1.
\end{aligned}
\end{equation} 
The overall sequential decision problem can then be written
\begin{equation}
\resizebox{0.48\textwidth}{!}{$\min_{c_k \in \mathcal{C} \setminus c_{1:k-1}} V(c_k; \theta_{\varpi_{c_{1:k-1}}}) = \min  \mathbb{E}_{c' \sim U(\mathcal{C})} \big[L(\theta_{\varpi_{c_k}}, c'; \theta_{\varpi_{c_{1:k-1}}})\big].$}
 \end{equation}

This formulation ensures that the estimator sequentially selects the most informative context to collect data in order to maximize overall performance across diverse environments. 

\section{Contextual NeuroMHE}

\begin{figure}
    \centering
    \includegraphics[width=1\linewidth]{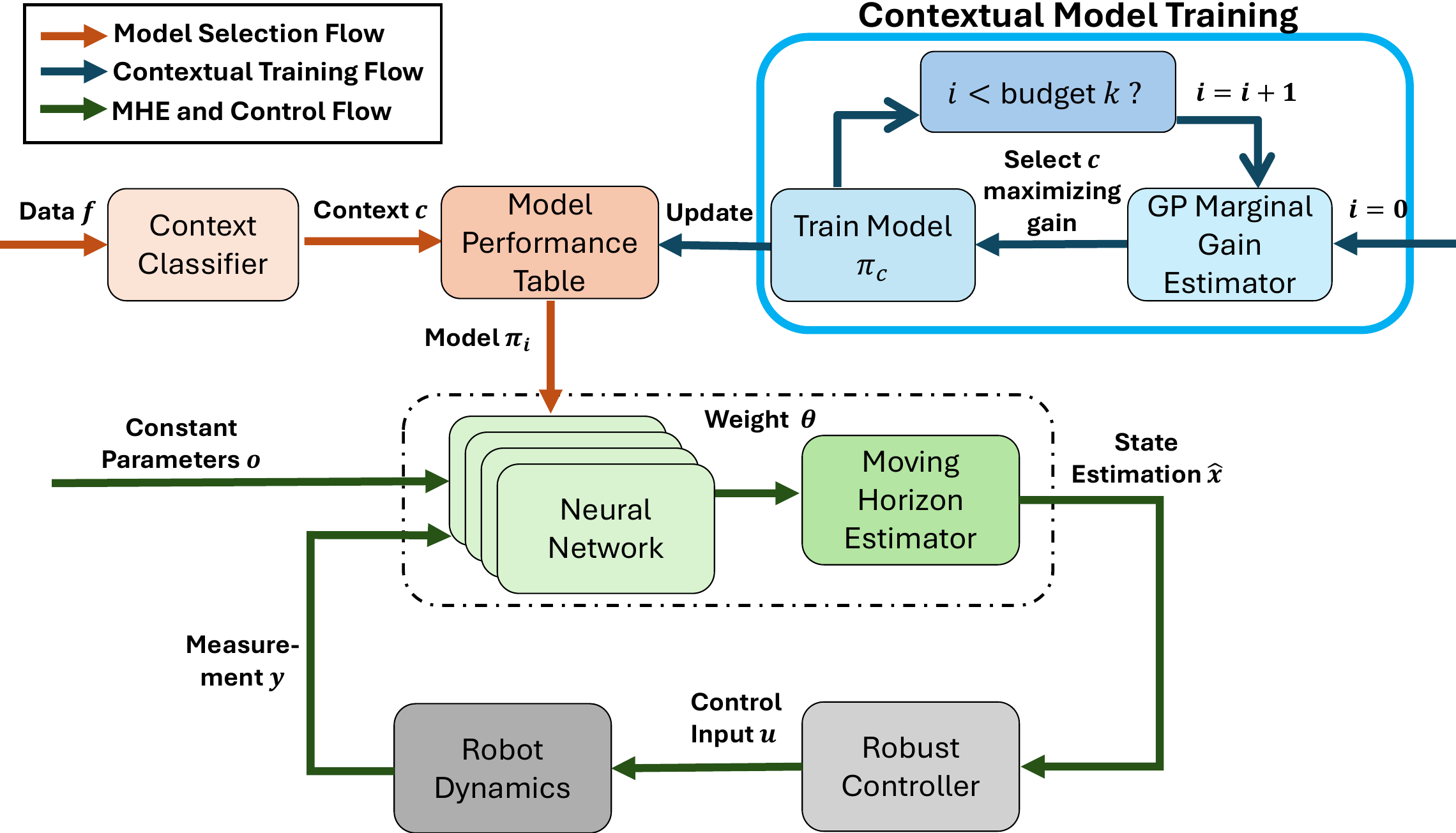}
    \caption{
        \textbf{System architecture for context-aware control with online model selection and adaptation.} (1) Training context selection via GP-based marginal gain estimation; (2) model selection of context-conditioned models $\pi_c$ with data $\mathcal{F}$; and (3) deployment using MHE-based state estimation $\hat{x}$ and robust control with context-specific weights $\theta$. The context $c$ is classified online and used to select the model and controller.
    }
    \label{fig:2}
    \vspace{-12 pt}
\end{figure}

\subsection{Overview}
Our approach consists of two main components: context selection and model training (Figure \ref{fig:2}). In the first stage, we use Bayesian Optimization with Gaussian Process regression to sequentially select the next most informative context from the remaining candidates, sequentially selecting each new context to train a corresponding model. At the end of this process, the trained models collectively form a representative training set that enables the drone to perform effectively across diverse scenarios. In the second stage, we train a NeuroMHE model for each selected context, learning context-dependent parameters that minimize estimation errors and improve generalization across the task space. Throughout this process, we maintain a value function $V$ to record and evaluate the performance of the sequential context selection strategy. The details of the algorithm are provided in Alg.~\ref{alg:1}.

\subsection{Assumption on Generalization Gap}

We consider the performance functions $J(\pi, c)$ and $V(c)$ for a given policy $\pi$ are assumed to vary smoothly over tasks. Our focus is on characterizing how performance degrades when transferring a policy trained on a source task to a different target task. We assume that the generalization gap between a source task $c$ and a target task $c'$ is modeled as a linear function of their distance:
\begin{equation}
    \Delta \hat{J}(\pi_c, c') = J(\pi_c, c) - \hat{J}(\pi_c, c') = \alpha \, \|c - c'\|,
\end{equation}
where $\alpha > 0$ is a slope parameter, and $\|c - c'\|$ denotes the distance between the task contexts $c$ and $c'$.

This assumption captures the intuition that the transferability of a policy decreases gradually with task dissimilarity. When two tasks are close in the task space, the generalization gap is small, indicating that the policy trained on the source task performs well on the target. As the distance increases, the gap grows linearly, reflecting a predictable decline in performance.

\subsection{Task Selection}

To determine the next context to train the model, we employ a Gaussian Process (GP) to model the performance across the contextual space. 
At each selection step $k$, a context $c_k$ is chosen by maximizing the acquisition function~\cite{cho2024model}:  

\begin{align}
    a(c; c_{1:k-1}) = \; \mathbb{E}_{c' \in \mathcal{X}} \Big[ 
    \big[ \mu_{k-1}(c) + \beta^{1/2} \sigma_{k-1}(c) 
    & \nonumber \\
    \; - \Delta \hat{J}(\pi_c, c') 
    - \hat{J}(\pi_{1:k-1}, c') \big]_+ \Big],
\end{align}
where $[\cdot]_+ = \max(\cdot, 0)$, and $\beta$ is a trade-off parameter between exploitation and exploration~\cite{cho2024model}.  

The performance values $\hat{J}(\pi_c, c)$ for all contexts not selected for training, $c \in \mathcal{X} \setminus {c_1, \ldots, c_k}$ are modeled using Gaussian Process (GP) regression. Let the selected training data up to iteration $k$ be denoted as $D_k = {(c_i, \hat{J}(\pi_{c_i}, c_i))}_{i=1}^{k}$. Given this data, the predicted GP posterior at a new context $c$ follows a normal distribution $P(\hat{J} \mid D_k) = \mathcal{N}(\mu_k(c), \sigma_k^2(c))$, where the predictive mean and variance are $\mu_k(c) = \mathbb{E}[\hat{J}(\pi_c, c)] + k^\top (K + \sigma^2 I)^{-1} y$ and $\sigma_k^2(c) = k(c, c) - k^\top (K + \sigma^2 I)^{-1} k$, respectively. Here, $k = [k(c, c_1), \ldots, k(c, c_k)]$ is the vector of covariances between $c$ and each observed context, and $K = [k(c_i, c_j)]{1 \leq i,j \leq k}$ is the covariance matrix over the observed inputs. As new observations are collected, the dataset $D_k$ is updated and the posterior distribution is refined, allowing the GP to progressively improve its estimate of the performance function.

\subsection{NeuroMHE}

The NeuroMHE framework~\cite{wang2023neural} is designed to learn context-dependent parameters that minimize state estimation errors while remaining computationally efficient. For each selected training context $c$, NeuroMHE learns the optimal parameter $\pi_{\varpi_c}$ by minimizing a context-specific loss function. The loss is chosen to directly penalize trajectory tracking errors and is defined as
\begin{equation}
    L(\hat{x}) = \sum_{k=t-N}^t \lVert \hat{x}_{k|t} - x_{k,\text{ref}} \rVert_W, 
\end{equation}
where $W \succ 0$ is a positive-definite weighting matrix, $\hat{x}_{k|t}$ is the estimated state at time $k$, and $x_{k,\text{ref}}$ denotes the corresponding reference trajectory. This formulation ensures that the estimator prioritizes accurate state tracking across the horizon.  

To optimize the parameters, the gradient of the loss with respect to $\pi$ is computed via the chain rule:  
\begin{equation}
    \tfrac{dL}{d\varpi} = \tfrac{\partial L}{\partial \hat{x}} \cdot \tfrac{\partial \hat{x}}{\partial \theta} \cdot \tfrac{\partial \theta}{\partial \varpi}.
\end{equation}
Here, the key term $\tfrac{\partial \hat{x}}{\partial \theta}$ is obtained using analytical gradients derived from a Kalman filter approximation~\cite{wang2023neural}. The state estimates $\{\hat{X}^{KF}_{k|k}\}^t_{k=t-N}$ are initialized as \begin{equation}
\begin{aligned}
P_{t-N} &= P^{-1},  C_{t-N} = (I - P_{t-N} S_{t-N})^{-1} P_{t-N}, \\
\hat{X}^{KF}_{t-N|t-N} &= (I + C_{t-N} S_{t-N}) \hat{X}_{t-N} + C_{t-N} T_{t-N}.
\end{aligned}
\end{equation}

The recursive estimation then proceeds for $k = t-N+1, \ldots, t$. At each step, the predicted state is updated by  
\begin{equation}
\hat{X}_{k|k-1} = \bar{F}_{k-1} \hat{X}^{KF}_{k-1|k-1} - G_{k-1} (L^{ww}_{k-1})^{-1}L^{w\theta}_{k-1},
\end{equation}
while the covariance and correction matrices are updated as  
\begin{equation}
P_k = \bar{F}_{k-1} C_{k-1} \bar{F}_{k-1}^\top + G_{k-1} (L^{ww}_{k-1})^{-1} G_{k-1}^\top,
\end{equation}
\begin{equation}
C_k = (I - P_k S_k)^{-1} P_k,
\end{equation}
and the filtered state is obtained by  
\begin{equation}
\hat{X}^{KF}_{k|k} = (I + C_k S_k) \hat{X}_{k|k-1} + C_k T_k.
\end{equation}

Finally, the dual variables are updated backward in time from $k = t$ to $t-N+1$, with $\Lambda^\ast_t = 0$, and the collection $\Lambda^\ast = \{\Lambda^\ast_{t-1}, \ldots, \Lambda^\ast_{t-N+1}\}$,
\begin{equation}
\begin{aligned}
\Lambda^\ast_{k-1} &= (I + S_k C_k)\,\bar{F}_{k}^\top \Lambda^\ast_{k+1} 
                 + S_k \hat{X}^{KF}_{k|k} + T_k.
\end{aligned}
\end{equation}
The optimal state estimates are then computed as  
\begin{equation}
\hat{X}_{k|t} = \hat{X}^{KF}_{k|k} + C_k \bar{F}_k^\top \Lambda^\ast_k.
\end{equation}
\begin{equation}
    \frac{\partial \hat{x}}{\partial \theta} = \{\hat{X}_{k|t}\}_{k=t-N}^{t}.
\end{equation}
This training procedure allows NeuroMHE to leverage analytical gradients for efficient optimization, ensuring stable and accurate state estimation across varying contexts.

\subsection{Weight Usage}
This section formalizes the role of the performance function in training and model selection. During training, after each model has been optimized on its assigned context, the performance function $V$ is updated to quantify the generalization capability of the resulting parameters. This update provides an evolving estimate of performance across the contextual space. At test time, no further training is performed; instead, model selection is governed by the maximization of the performance function: $\varpi^{\star} \in \arg\max_{\varpi} V(c; \varpi)$, which yields the parameterization that achieves the highest expected performance for a given context $c$. This corresponds directly to the Model Performance Table illustrated in Figure~\ref{fig:2}, where $V$ records the performance of each candidate model across contexts and guides the selection of the best-performing network. In this way, the deployed estimator is chosen systematically according to demonstrated generalization performance.

\begin{algorithm}[t]
\caption{Contextual Learning for Task Training with Model-based Policy Gradient and MHE}
\textbf{Input:} Task (context) set $ X $, Training budget $ K $ \\
\textbf{Initialization:} $ J, V = 0 \, \forall x \in X $, $ \boldsymbol{\varpi} = \emptyset $, $ k = 1 $
\label{alg:1}
\begin{algorithmic}[1]
\While{ $ k \leq K $ }
    \State \% Estimate training performance
    \State $ \mu, \sigma \gets \text{GP}(E[J(\theta(\varpi), c)], k(c, \tilde{c})) $
    \State \% Calculate marginal generalized performance and acquisition function
    \State Calculate $ a(c; x_{1:c-1}) $ with Eq. 7
    \State \% Select the next training task
    \State $ c_k = \arg\max_{c} a(c; x_{1:c-1}) $
    
    \State \% Train using Policy Gradient with MHE
    \While{$ L_\text{mean}$ does not converged}
        \For{$t = 0$ to $ T_{\text{episode}} $}
            \State \textbf{Forward Pass:}
            \State Compute adaptive weights $\theta^t$ using the neural network parameters $\varpi^t_k$.
            \State Estimate $\hat{x}$ by solving MHE in Eq.~(4).
            \State Compute control input $u^t$.
            \State Apply $u^t$ to update the quadrotor state $x^t$.
            
            \State \textbf{Backward Pass:}
            \State Compute $\frac{\partial \hat{x}}{\partial \theta}$ using the Kalman filter approximation Eqs.~(10--16).
            \State Compute $\frac{\partial L}{\partial \varpi}$ using Eq.~(9).
            \State Update network parameters $\varpi^t_k$.
            
        \EndFor
        \State Calculate $ L_{\text{mean}} $ for the next episode 
    \EndWhile
    \State $ \boldsymbol{\varpi} \gets \boldsymbol{\varpi} \cup \{\varpi_k^t\} $
    \State $ k \gets k + 1 $
\EndWhile
\State \% calculate generalization performance $ V(\varpi_1, \dots, \varpi_K) $
\State \textbf{Output:} Set of policies $ \boldsymbol{\varpi} $ and performance $ V $
\end{algorithmic}
\end{algorithm}

\section{Real-World Evaluation}
\begin{figure}[t]
    \centering
    \includegraphics[width=1.0\linewidth]{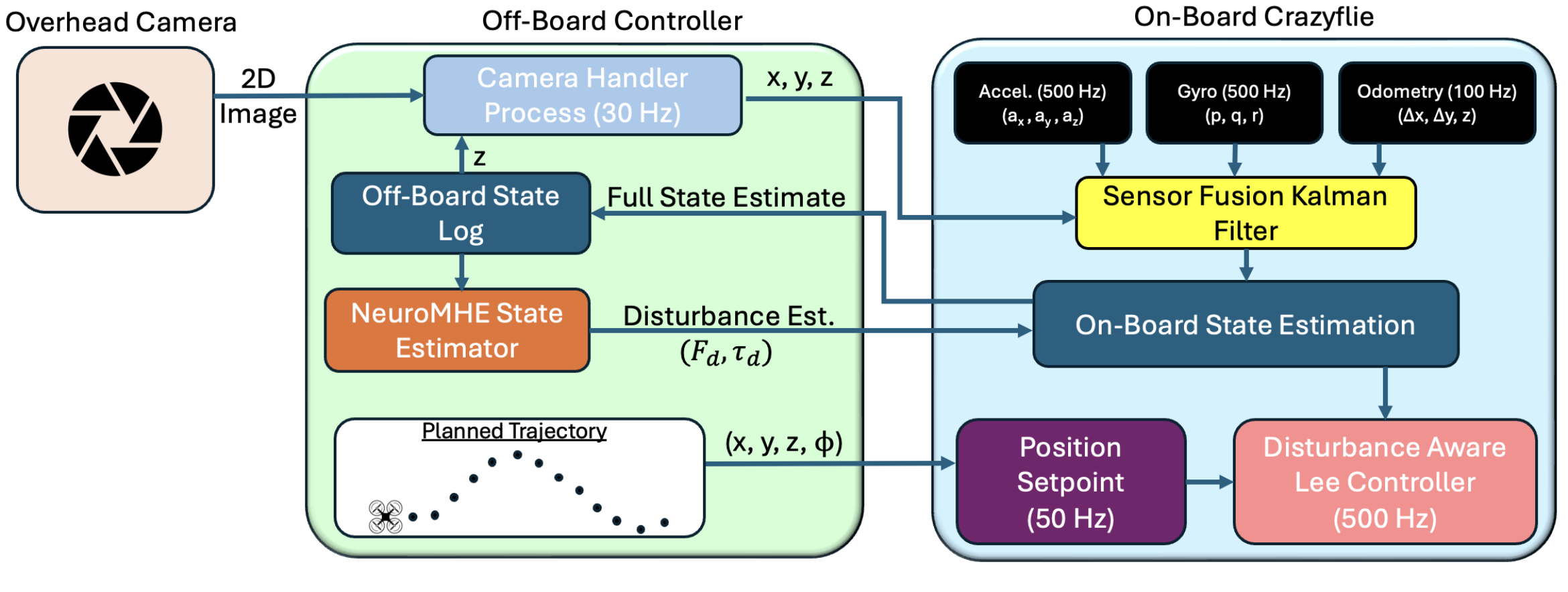}
    \caption{The system diagram. z-position is computed using the FlowV2 odometry sensor.}
    \label{fig:process_setup}
    \vspace{-10 pt}
\end{figure}

We evaluated our method against multiple baselines in real-world experiments using a Crazyflie2.1 and multiple common, household fans. We only performed real-world experiments because airflow simulators often fail to capture near-field turbulence, ground effect, and prop–airframe interactions \cite{Oo2023TurbulenceUAV} \cite{Coursey2024SimToRealGap}. The following subsections detail our setup. 

\subsection{Training Data}
We collected 13 contexts of varying wind speeds and directions. One context had \textit{No Wind}, while the remaining 12 contexts were combinations of six wind directions (\textit{\{Left Crosswind, Right Crosswind, Headwind, Tailwind, Updraft, Downdraft\}}) and two wind speeds (\textit{\{Low, High\}}). For each context, we collected five samples of the drone traveling in a straight line through the wind flow.  

\subsection{Disturbance-Aware Lee Controller}
Since NeuroMHE outputs estimates of $F_{\text{dist}}$ and $\tau_{\text{dist}}$, we enhanced the Lee Controller \cite{lee2011controlcomplexmaneuversquadrotor} to take into account these two additional values when calculating desired thrust $f$ and the moment vector $M$:
\begin{equation}
f=\left(k_x \mathbf{e}_x + k_v \mathbf{e}_v + m g\, \mathbf{e}_3 - m \ddot{\mathbf{x}}_d - F_{dist}\right)\cdot R \mathbf{e}_3 
\end{equation}
\begin{equation}
\begin{aligned}
M = -k_R e_R - k_{\Omega} e_{\Omega}
    + \Omega \times (J\,\Omega) \\
    - J\!\left(\hat{\Omega}\, R^{\top} R_c\, \Omega_c
               - R^{\top} R_c\, \dot{\Omega}_c \right) - \tau_{dist}.
\end{aligned}
\end{equation}
Here, $m$ is the drone's mass, $J \in \mathbb{R}^{3x3}$ is the inertia matrix with respect to the
body-fixed frame, $R \in SO(3)$ is the rotation matrix with respect to the body frame, $\Omega$ is the angular velocity respect to the body-frame, $\ddot{x}_d$ is the linear acceleration with respect to the world frame, the $e$ terms are error terms, and $k$ terms are tuned positive constants.

Then, we obtain the thrusts for each individual motor, $f_1, f_2, f_3, f_4$ through: 
\begin{equation}
\begin{aligned}
\begin{bmatrix}
f_1\\
f_2\\
f_3\\
f_4
\end{bmatrix}
=
\begin{bmatrix}
1 & 1 & 1 & 1\\
0 & -d & 0 & d\\
d & 0 & -d & 0\\
- c_{\tau} & c_{\tau} & - c_{\tau} & c_{\tau}
\end{bmatrix}^{-1}
\begin{bmatrix}
f\\
M_1\\
M_2\\
M_3
\end{bmatrix}
\end{aligned}
\end{equation}
where $d$ is the distance from each of the motors to the center of the drone, and $c_\tau$ is a fixed constant. 

\begin{figure}[h]
  \centering
  \subfloat[Environment One]{\includegraphics[width=0.32\linewidth, 
  height=0.3\linewidth]{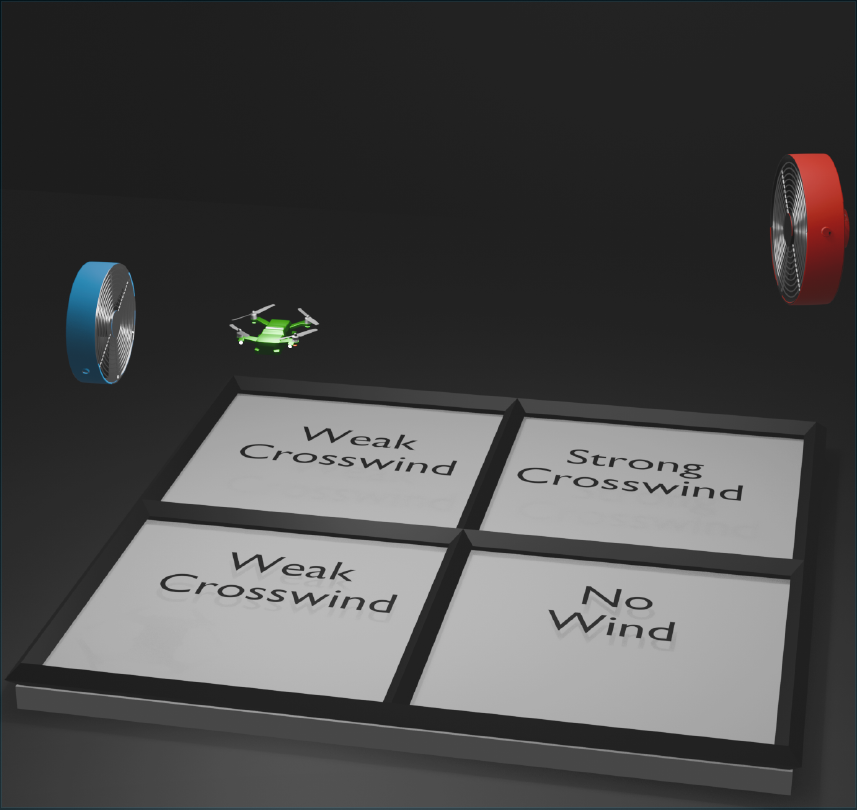}\label{fig:a}}\hfill
  \subfloat[Environment Two]{\includegraphics[width=0.32\linewidth,
  height=0.3\linewidth]{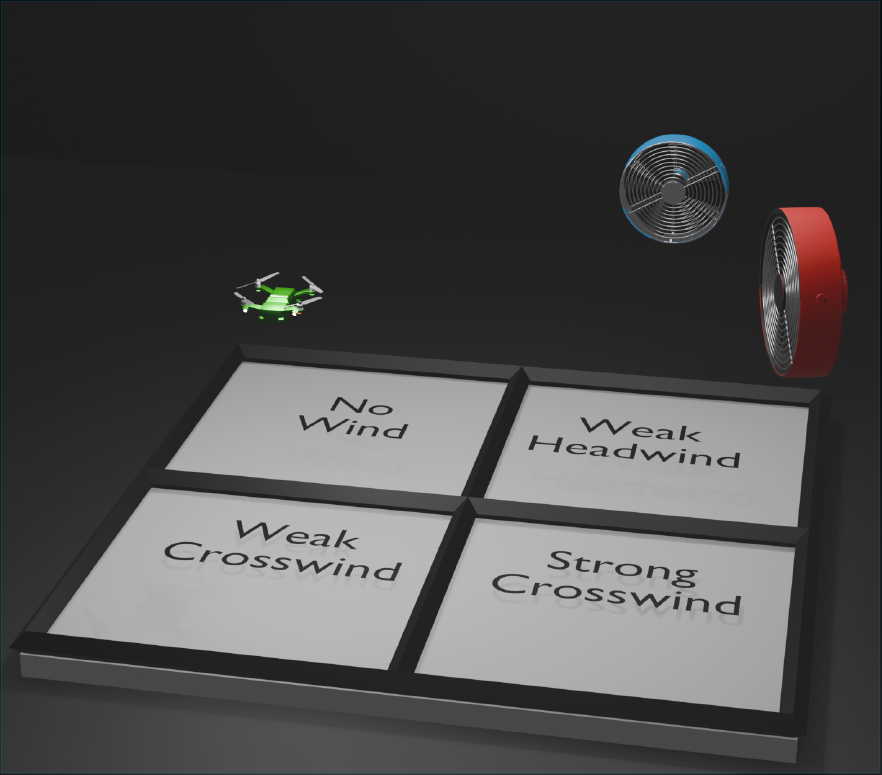}\label{fig:b}}\hfill
  \subfloat[Environment Three]{\includegraphics[width=0.32\linewidth,
  height=0.3\linewidth]{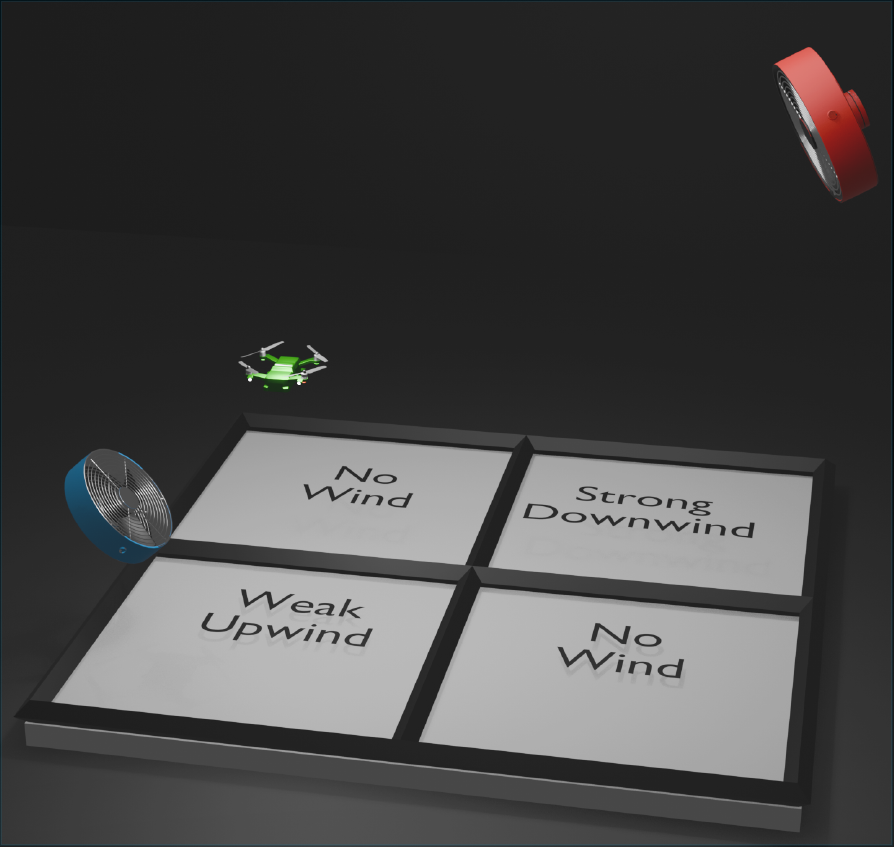}\label{fig:c}}
  \caption{3D illustration of our three test environments. Each environment has a $1.5m$ x $1.5m$ x $1m$ designated flight space. The red fan represents the \textit{strong} fan, while the blue fan represents the \textit{weak} fan. The specific context is written on the ground of the four quadrants.}
  \label{fig:environments}
\end{figure}

\subsection{System Parameters}
The neural networks are implemented as fully connected feedforward models with two hidden layers of 30 units each and ReLU activations. Each network takes an input of size six and produces a 25-dimensional output corresponding to the flattened parameterization of the MHE weighting matrices. The notion of context $c \in \mathbb{R}^2$ (wind direction and wind speed provided to GP) is used to define training scenarios but is not directly provided as input to the network. Training is performed using the Adam optimizer with a learning rate of $10^{-4}$, and each model is trained until the loss difference between successive epochs falls below $10^{-3}$. For GP-based context selection, we employ an RBF kernel with length scale $1.0$ and set the trade-off parameter to $\beta = 1.0$ with slope parameter $10^{-3}$ to balance exploration and exploitation.

\subsection{Physical Setup}

Our physical setup featured an off-board computer, a Crazyflie drone, and an overhead camera (Figure \ref{fig:process_setup}). Due to limited computation on the Crazyflie, we ran the GP and NeuroMHE models off-board on a laptop equipped with a 12-core Intel i9-8950HK CPU and RTX 2080 Mobile. The drone computed its state using four sensors: an accelerometer, a gyroscope, an odometry sensor, and an overhead camera. Path following was accomplished through updating the position setpoint $(x, y, z, \phi)$ at 50Hz.


\begin{figure}[t]
  \centering
  \subfloat[Hover Trajectories]{\includegraphics[width=0.33\linewidth]{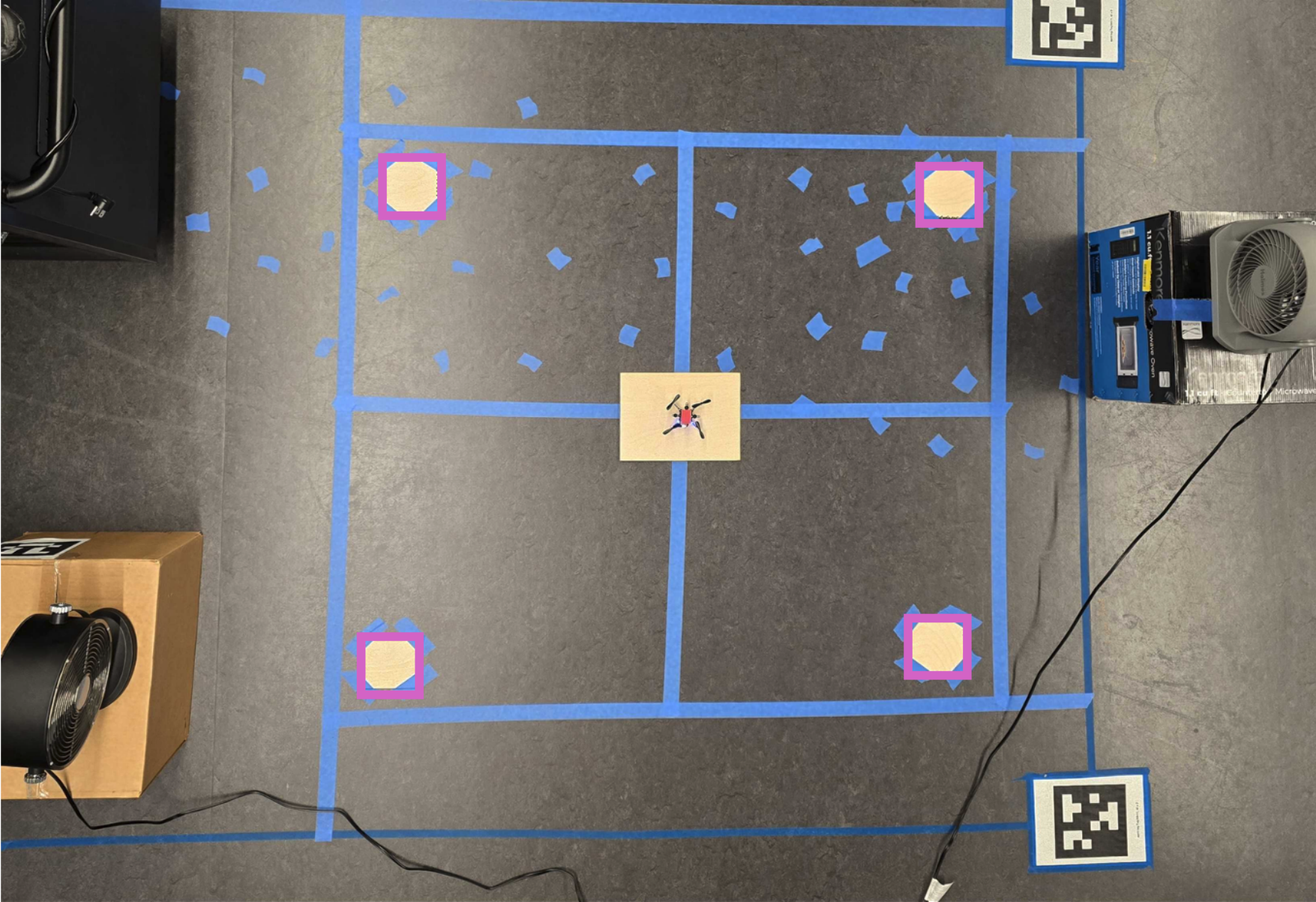}\label{fig:hover}}\hfill
  \subfloat[Square Trajectory]{\includegraphics[width=0.33\linewidth]{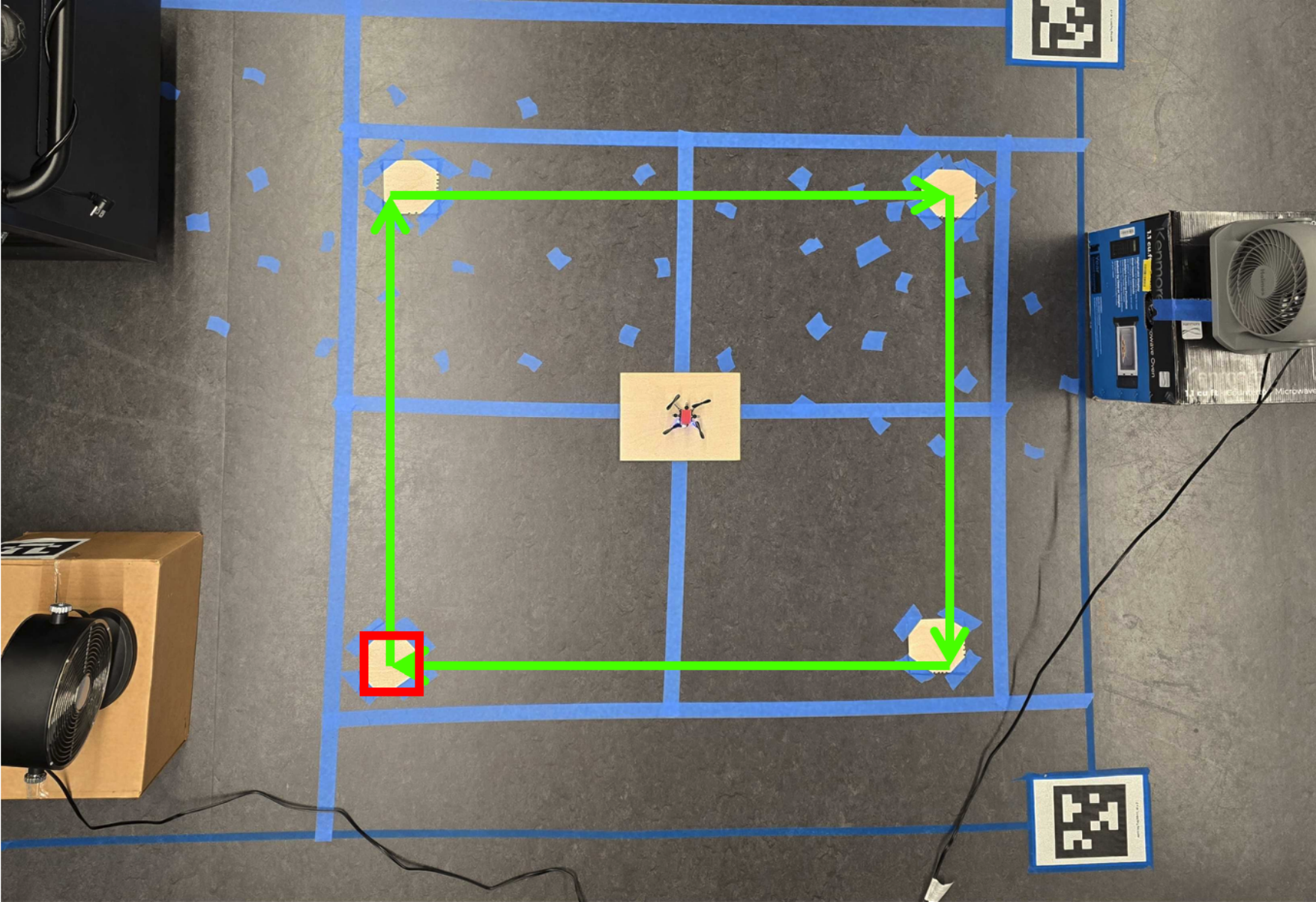}\label{fig:square}}\hfill
  \subfloat[Figure-8 Trajectory]{\includegraphics[width=0.33\linewidth]{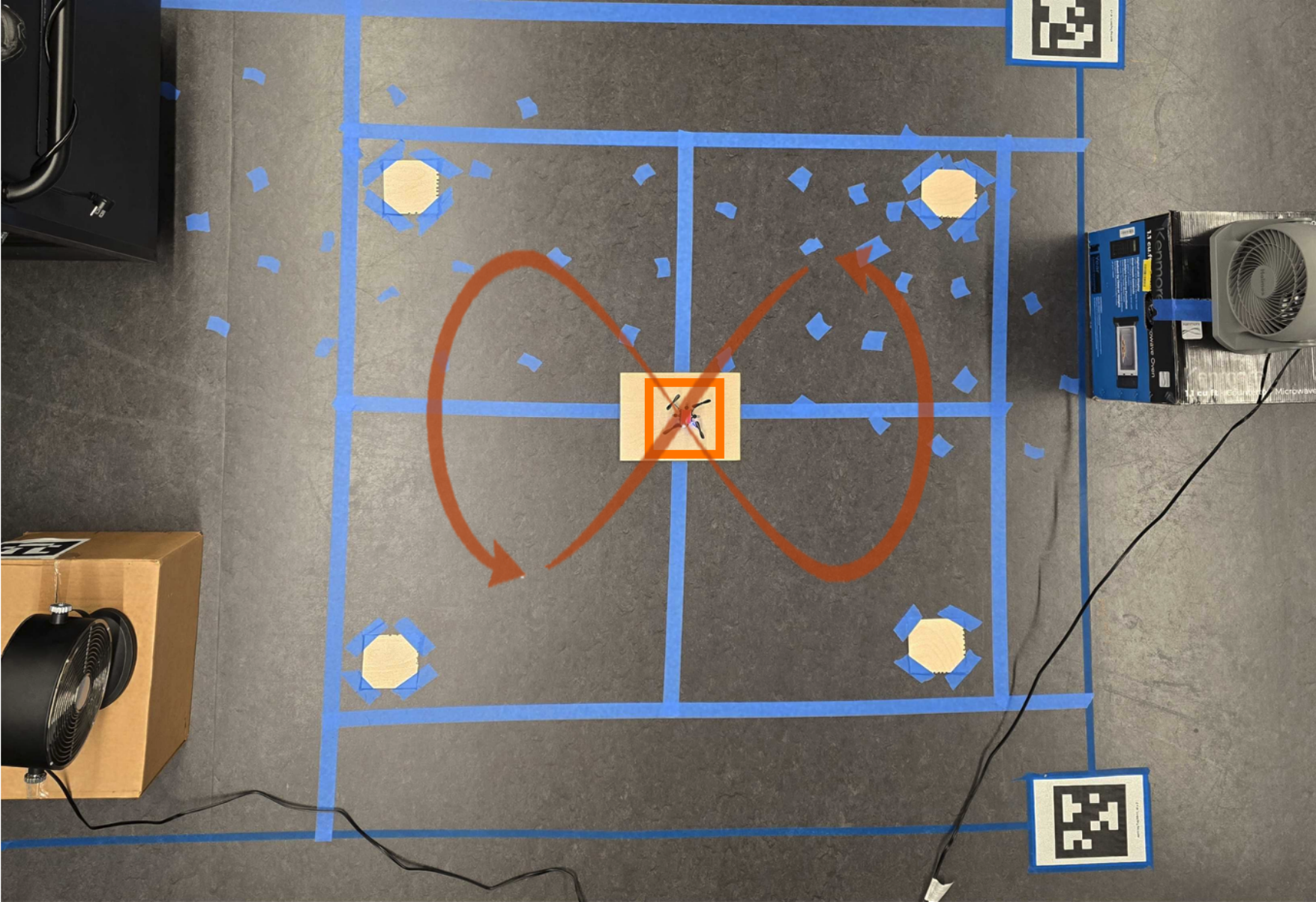}\label{fig:figure-8}}
  \caption{ Bird's eye view of our experimental trajectories overlaid atop the actual physical setup. We placed two fans (the grey fan is the \textit{strong} fan and the black one is the \textit{weak} fan) on top of $0.5m$ high boxes for the first two environments. In the third environment, the \textit{weak} fan is placed on the ground and the \textit{strong} fan is placed on top of a taller box, each angled at $45\degree$.}
  \label{fig:environments}
  \vspace{-10 pt}
\end{figure}

\begin{figure}[t]
    \centering
    \includegraphics[width=1.00\linewidth]{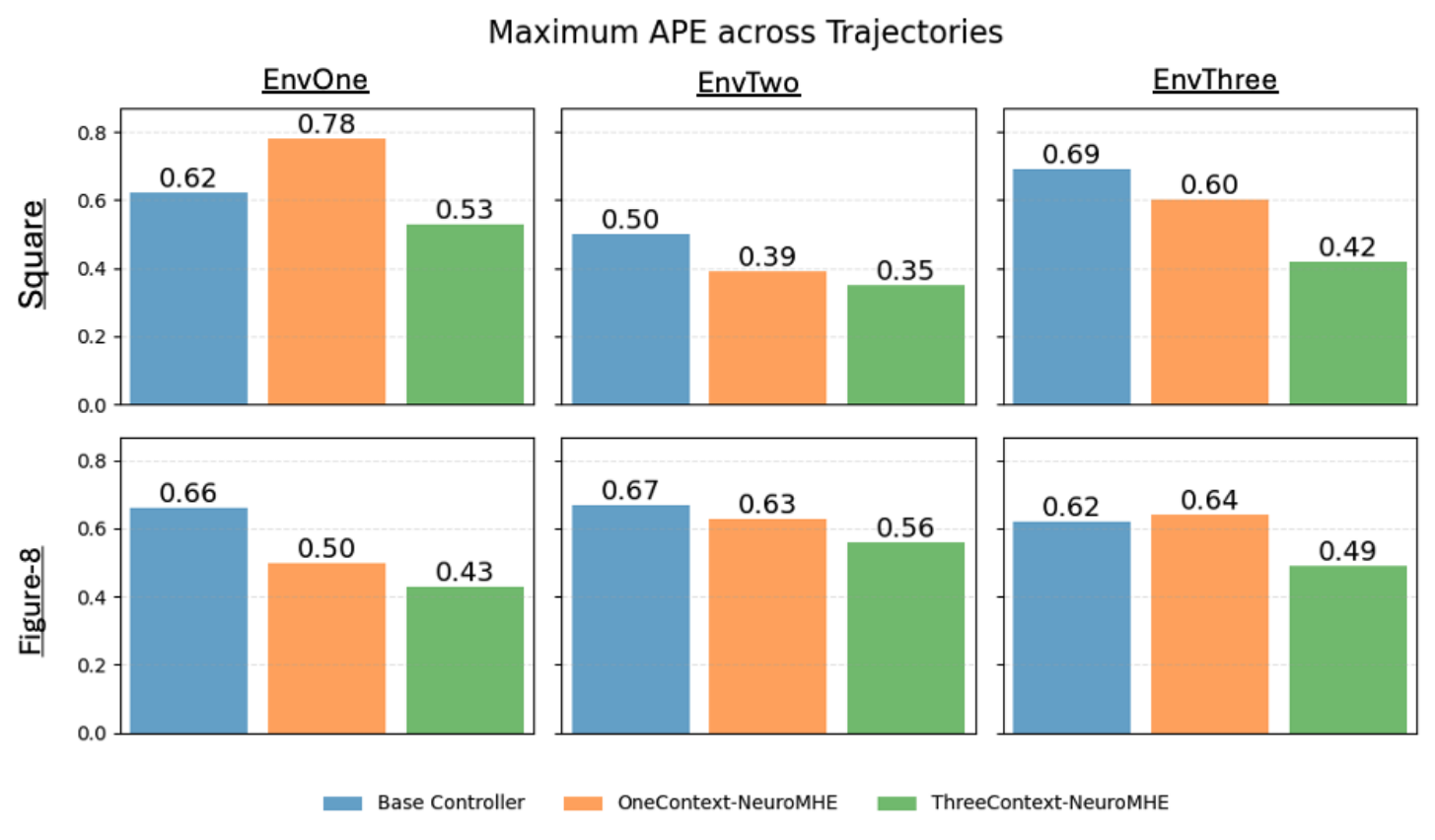}
    \caption{The maximum APE, for the square and figure-8 trajectories, which is defined as the maximum offset of the recorded state from the desired setpoint across all steps.}
    \label{fig:max_ape}
\end{figure}
\subsection{Testing Trajectories}
We defined three different trajectories for experimentation.

\textbf{Hover}: Starting in any of the four corners (magenta squares in Figure \ref{fig:hover}), the drone rose $0.5m$, held that position for five seconds, then landed back in the starting position. 
    
\textbf{Square}: Starting in the bottom left corner (red square in Figure \ref{fig:square}), the drone rose $0.5m$, then visited each of the four corners clockwise. The drone traveled at $0.3m/s$ throughout the trajectory. 
    
\textbf{Figure-8}: Starting from the center of the environment (orange square in Figure \ref{fig:figure-8}), the drone rose $0.5m$ and did a figure-8 trajectory on the $xy$ plane at $0.3m/s$. 

\begin{figure*}[t]
    \centering
    \includegraphics[width=0.85\linewidth]{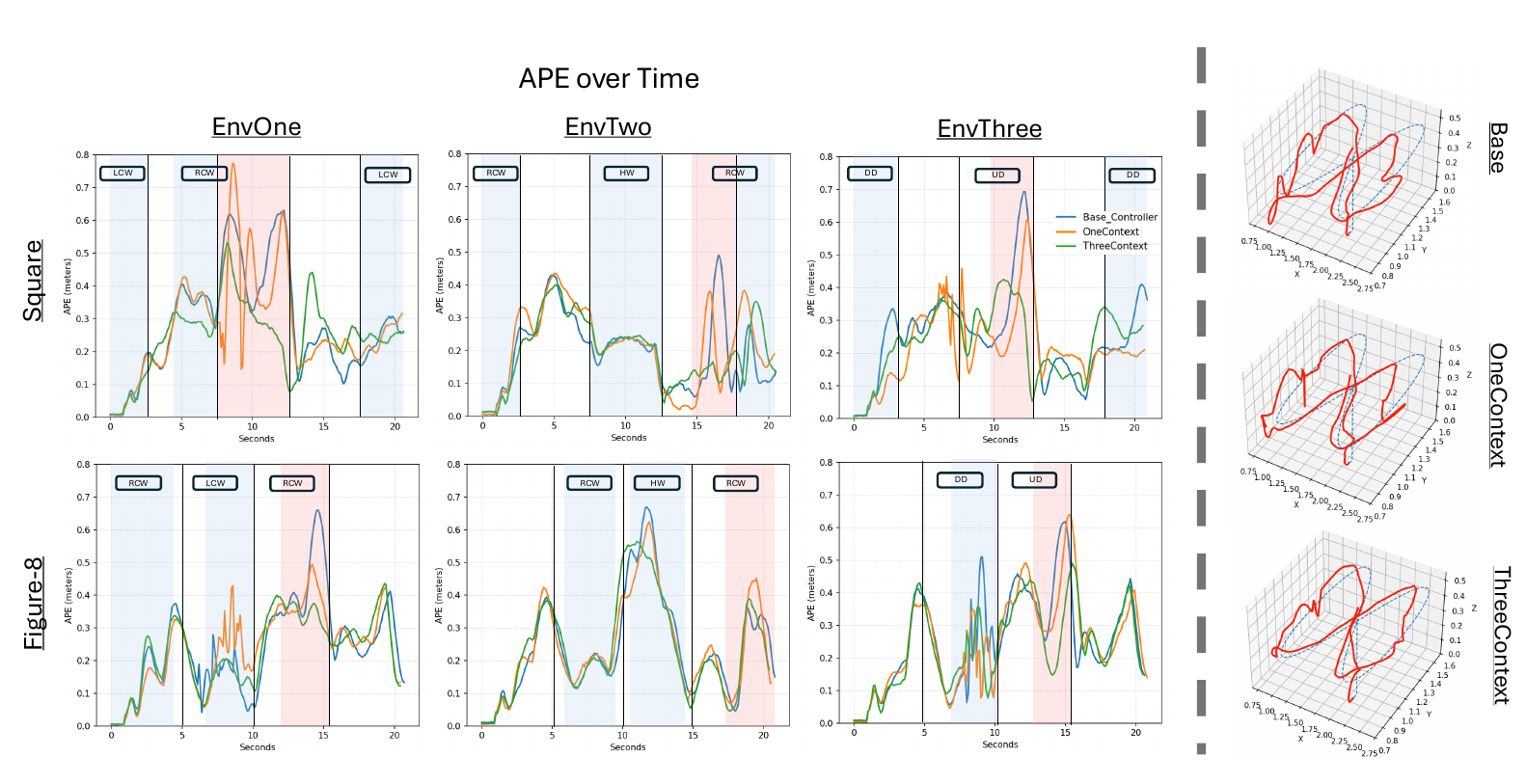}
    \caption{(Left) Absolute Position Error (APE) for the square and figure-8 trajectories in three environments. The blue shaded regions indicate a low-wind flow, the red-shaded regions indicate a high-wind flow, and the black vertical lines indicate a context switch. The wind flows may not cover the entire context, causing the occasional discrepancy between the context switch and the indicated wind flow. (Right) 3D Plots of the Figure-8 Trajectory flown in EnvThree (solid red is the actual path and dotted blue is the reference). LCW/RCW - Left/Right Crosswind, HW - Headwind, UD/DD - Up/Down Draft.}
    \label{fig:ape_over_time}
\end{figure*}

\begin{table*}[t]
\centering
\caption{RMSE Absolute Position Error (APE) across environments and trajectories. The different hover trajectories are aggregated together. With just three contexts, we perform comparably with the oracle and much better than the baseline.}
\setlength{\tabcolsep}{6pt}
\begin{tabular}{l *{9}{c}}
\toprule
& \multicolumn{3}{c}{Environment One} & \multicolumn{3}{c}{Environment Two} & \multicolumn{3}{c}{Environment Three}  \\

\cmidrule(lr){2-4} \cmidrule(lr){5-7} \cmidrule(lr){8-10}
Method & Hover & Square & Figure-8 & Hover & Square & Figure-8 & Hover & Square & Figure-8 \\
\midrule
Base Controller       & 0.4524 & 0.2743 & 0.4119 & 0.1612 & 0.2256 & 0.3560 & \textbf{0.1249} & 0.2887 & 0.3446 \\
OneContext-NeuroMHE   & 0.1679 & 0.3174 & 0.3966 & 0.1650 & 0.2435 & \textbf{0.3363} & 0.1483 & \textbf{0.2313} & 0.3553 \\
ThreeContext-NeuroMHE & \textbf{0.1379} & \textbf{0.2693} & \textbf{0.3896} & \textbf{0.1528} & \textbf{0.2218} & 0.3397 & 0.1479 & 0.2561 & \textbf{0.3305} \\
\rowcolor{gray!30}
FullContext-NeuroMHE & \textbf{0.1240} & \textbf{0.2622} & \textbf{0.3641} & \textbf{0.1344} & \textbf{0.2036} & \textbf{0.3344} & \textbf{0.1205} & 0.2467 & \textbf{0.3207} \\
\bottomrule
\end{tabular}
\label{table:table}
\end{table*}

\subsection{Evaluated Controllers}
\textbf{Base Controller}: A standard Extended Kalman Filter for state estimation and a standard Lee Controller (Disturbance-Unaware) for low-level control. 

\textbf{OneContext-NeuroMHE}: A single NeuroMHE model used for all contexts coupled with a Disturbance-Aware Lee Controller for low-level control. We evaluated the performance of all 13 models (one model for each context) and picked the ``No-Wind'' model since it had the highest general performance. This model is essentially equivalent to the original NeuroMHE \cite{wang2023neural}.

\textbf{ThreeContext-NeuroMHE}: Our Contextual NeuroMHE trained with a budget of three.
    
\textbf{FullContext-NeuroMHE}: We trained a NeuroMHE model for each of the 13 contexts and switched to the matching model whenever that context was active, yielding a full oracle-style controller that served as our lower bound.

\subsection{Results}

From our experiments, the ThreeContext-NeuroMHE had a $14.9\%$ and a $4.9\%$ reduction in absolute tracking error on average over the Base Controller and OneContext-NeuroMHE, respectively. (See Table \ref{table:table}) The major strength of context switching occurs when traveling through abrupt changes in wind disturbances. 

\vspace{-1pt}

Since the drone deviates most during these disturbance shifts, we examine the maximum APE of the square and figure-8 trajectories. Referring to the six evaluated cases in Figure \ref{fig:max_ape}, the Base Controller, OneContext-NeuroMHE, and ThreeContext-NeuroMHE had average Maximum APE scores of $0.62$, $0.59$, and $0.47$, respectively. Therefore, the ThreeContext-NeuroMHE achieves a $24.2\%$ and \textbf{$20.3\%$} decrease in Maximum APE over the Base Controller and OneContext-NeuroMHE, respectively. This decrease in tracking error can be attributed to the context-specific models having a much stronger prior in those specific wind-conditions, needing less time for the MHE to provide accurate state and disturbance estimations, further showcasing the benefit of enabling context switching. 

However, due to the increased communication overhead necessary for Contextual NeuroMHE, there is occasional packet loss, causing occasional drifts around tight turns. A more capable drone would be able to run all processes on board, which would circumvent this issue.

\section{Discussion and Conclusion}
 Contextual NeuroMHE adapts to abrupt wind disturbances much more effectively than NeuroMHE and our EKF baseline. We highlight that even with a highly constrained Gaussian Process training budget, our method achieves a $20.3\%$ decrease in absolute tracking error, which can be attributed to context-specific models having stronger priors in these new settings, requiring fewer control steps to converge. 

While our proposed approach demonstrates promising results in real-world experiments, there are several directions for improvement. First, the current framework primarily relies on offline learning, which makes it susceptible to out-of-distribution disturbances not captured during training. Extending the method toward online or continual learning would enable the quadrotor to adapt more effectively to novel and rapidly changing conditions. Second, the backward pass computation, which leverages Kalman filter approximations for estimating derivatives, remains computationally intensive. Future research could explore simplified or approximate formulations that reduce computational overhead while preserving accuracy.

\bibliographystyle{IEEEtran}
\bibliography{IEEEexample}

\end{document}